\pdfoutput=1

\documentclass[11pt]{article}

\usepackage[preprint]{acl}

\usepackage{times}
\usepackage{latexsym}

\usepackage[T1]{fontenc}

\usepackage[utf8]{inputenc}

\usepackage{microtype}

\usepackage{inconsolata}

\usepackage{graphicx}

\newcommand{\citeit}[1]{\textcolor{orange}{\textit{[CITE]}}}

\newcommand{\anonreview}[1]{}

\newcommand{\ignore}[1]{}
\newcommand{\nop}[1]{}

\newcommand{\eg}{\textit{e.g.,~}}
\newcommand{\ie}{\textit{i.e.,~}}

\newcommand{\baselineGPT}{Baseline GPT2~}
\newcommand{\baselineBLOOM}{Baseline BLOOM~}
\newcommand{\aisleGPT}{AISLE$_{GPT2}$~} 
\newcommand{\aisleBLOOM}{AISLE$_{BLOOM}$~}

\newcommand{\domainpretrained}{domain pre-trained}

\newcommand{\chemner}{CHEMDNER}

\newcommand{\finetune}{fine-tune~}
\newcommand{\finetuning}{fine-tuning~}
\newcommand{\finetuned}{fine-tuned~}
\newcommand{\Finetuning}{Fine-tuning~} 
\newcommand{\Finetuned}{Fine-tuned~}

\newcommand{\offtheshelf}{off-the-shelf~}
  \usepackage{numprint} 

\usepackage{tikz} 
\usepackage{pgfplots} 
\usepackage{pgfplotstable} 
\usepackage{csvsimple}
\usepackage{pgf-pie}
\usepackage{comment}
\usepackage{amsmath}

 \definecolor{DARKPURPLE}{rgb}{0.3194,0.1343,0.5522}
 \definecolor{aisleGPT}{rgb}{0.3947,0.2776,0.6201}
 \definecolor{aisleBloom}{rgb}{0.5661,0.5522,0.7593}
 
 \definecolor{baseGPT}{rgb}{0.1507,0.4645,0.7208}
 \definecolor{baseBloom}{rgb}{0.5357,0.7461,0.8643}

 \definecolor{WosColor}{rgb}{0.0314,0.2816,0.5583}
 \definecolor{DblpColor}{rgb}{0.0748,0.3733,0.6552}
 \definecolor{MagColor}{rgb}{0.1507,0.4645,0.7208}
 \definecolor{AaaiColor}{rgb}{0.2482,0.5619,0.771}
 \definecolor{AclColor}{rgb}{0.3616,0.6427,0.8166}
 \definecolor{IclrColor}{rgb}{0.4918,0.722,0.8548}
 \definecolor{IcmlColor}{rgb}{0.6325,0.7976,0.8869}
 \definecolor{NeuripsColor}{rgb}{0.7506,0.8478,0.9282}
 \definecolor{ScopusColor}{rgb}{0.8289,0.8938,0.9547}
 \definecolor{PubmedColor}{rgb}{0.5357,0.7461,0.8643}

 \definecolor{OstiColor}{rgb}{0.3194,0.1343,0.5522}
 \definecolor{AminerColor}{rgb}{0.3947,0.2776,0.6201}
 \definecolor{Cord19Color}{rgb}{0.4732,0.4327,0.6993}
 \definecolor{CoreColor}{rgb}{0.5661,0.5522,0.7593}
 \definecolor{S2OrcColor}{rgb}{0.6731,0.6664,0.82}
 \definecolor{GoogleScholarColor}{rgb}{0.7765,0.7791,0.8824}
 \definecolor{Elibrary.RuColor}{rgb}{0.8749,0.873,0.9311}

 \definecolor{ArxivColor}{rgb}{0.7365,0.08,0.1012}
 \definecolor{BiorxivColor}{rgb}{0.9467,0.2682,0.1961}
 \definecolor{ArabixivColor}{rgb}{0.9875,0.5412,0.4157}
   
 \definecolor{topkvenue}{RGB}{9, 121, 105}
 \definecolor{venuelisted}{RGB}{42, 170, 138}
 \definecolor{venuemissing}{RGB}{248, 200, 220}

\usepackage{multirow}
\usepackage{array}
\newcolumntype{H}{>{\setbox0=\hbox\bgroup}c<{\egroup}@{}}

\usepackage{colortbl, xcolor}

\usepackage[none]{hyphenat}

\usepackage{lipsum}
\usepackage{enumitem}
\usepackage{multirow}
\usepackage{booktabs}
\usepackage{xcolor}

\title{Exploring the Benefits of Domain-Pretraining of Generative Large Language Models for Chemistry}

\author{Anurag Acharya, ~~Shivam Sharma, ~~Robin Cosbey, ~~Megha Subramanian \\
        \textbf{Scott Howland}, ~~\textbf{Maria Glenski} \\
        Pacific Northwest National Laboratory\\
        Richland, WA 99354 \\
        \texttt{\{anurag.acharya, shivam.sharma, robin.cosbey, megha.subramanian,} \\
        \texttt{scott.howland, maria.glenski\}@pnnl.gov}
        }        

\begin{document}
\maketitle
\begin{abstract}
A proliferation of Large Language Models (the GPT series, BLOOM, LLaMA, and more) are driving forward novel development of multipurpose AI for a variety of tasks, particularly natural language processing (NLP) tasks. These models demonstrate strong performance on a range of tasks; however, there has been evidence of brittleness when applied to more niche or narrow domains where hallucinations or fluent but incorrect responses reduce performance. Given the complex nature of scientific domains, it is prudent to investigate the trade-offs of leveraging \offtheshelf versus more targeted foundation models for scientific domains. In this work, we examine the benefits of in-domain pre-training for a given scientific domain, chemistry, and compare these to open-source, off-the-shelf models with zero-shot and few-shot prompting. Our results show that not only do in-domain base models perform reasonably well on in-domain tasks in a zero-shot setting but that further adaptation using instruction \finetuning yields impressive performance on chemistry-specific tasks such as named entity recognition and molecular formula generation.
\end{abstract}

\newcommand{\fix}{\marginpar{FIX}}
\newcommand{\new}{\marginpar{NEW}}


\section{Introduction}
Large Language Models (LLMs) and foundation models are becoming increasingly ubiquitous~\cite{bommasani2021opportunities} across  traditional natural language processing applications~\cite{aharoni2019massively, brown2020language} as well as a wide array of domains that can leverage natural language knowledge, reasoning, or enrichments (\eg law, medicine, biology). Architectures leveraging self-supervised training at scale have become more common as well~\cite{ranftl2021vision}. 
This has resulted in a vast increase in use of LLMs, and more broadly AI, across a variety of domains from commonsense reasoning to law and from natural sciences to cultural motifs (\citealt{sap2019atomic}, \citealt{kell2020deep}, \citealt{acharya2021atlas}, \citealt{yarlott2021finding}, \citealt{gu2021domain}, \citealt{xiao2021lawformer}, \citealt{moor2023foundation}).

While foundation models have displayed a strong capability to generalize to unseen tasks and to perform in-context reasoning, they struggle to produce consistent, \textit{factual} output on many tasks~\cite{xiao2021hallucination}. This weakness is of particular importance in various scientific or otherwise technical domains, where factual incoherence is a critical impediment to adoption or impactful use where they could otherwise prove incredibly beneficial to domain experts and practitioners as a means of augmenting human expertise for accelerated advances.

Due to the vast compute required to train or tune a model at scale, it is critical to understand the trade-off of different strategies to boost performance in domain whether through pre-training from scratch on domain-rich pre-training data, approaches to fine-tune a pre-trained model, or stacking self-supervised pre-training and task-focused fine-tuning. In addition, there is an increasing reliance on interactions with these foundation models using API access (where requests are sent to server-backed instantiations rather than processed locally). 

Reliance on state-of-the-art models via API access may raise concerns or impede adoption of these methods -- \eg cost concerns (cost of API requests), privacy concerns (information contained within the request being shared externally), and memorization impacts (a subsequent query of the model hitting a version that has been retrained or updated using the previous requests for settings where that would be undesirable). As these (non-empirical) trade-offs that may also motivate a potential choice to leverage the technology of foundation models at a smaller scale for heightened control (on what data models are trained on, what architectures are used, or what scale the models are trained to) increase in consideration, it is important to evaluate performance of both local instantiation strategies and \offtheshelf models.

In this work, we focus on a specific scientific field -- chemistry -- and, leveraging scientific literature from a multitude of sources, we curate a large and diverse chemistry publication dataset to train several chemistry-focused foundational models. We introduce these foundational models for chemistry as AISLE (AI from Scientific Literature) models and present an analysis that quantifies the benefit of in-domain pre-training or \finetuning compared to \offtheshelf baselines. Our experiments analyze the benefit of in-domain pre-training from scratch and/or task \finetuning leveraging instruction based prompts for domain-specific tasks with general benchmarking provided for context.

Additionally, we adapt the models further by instruction \finetuning on some chemistry-specific tasks. We see that the performance of models instruction \finetuned on a combination of all these tasks perform exceptionally well, indicating that is not necessary to perform instruction \finetuning for each individual task to adapt a model to a domain. 
We also observe some limitations of these systems, suggesting a potential need to expose the models to perhaps an even more diverse collection of data in the future, \eg going beyond  scientific literature to structural representations of molecular structure and other chemical properties. Overall, this work introduces the following novel contributions:

\begin{enumerate}
    \setlength\itemsep{-0.5em}
    \item we introduce AISLE models, a set of novel chemistry foundation models pre-trained from scratch on chemistry-focused scientific literature;
    \item an analysis of performance gains from domain-adapted models pre-trained from scratch in zero- and few-shot settings; and
    \item an empirical comparison of \offtheshelf models compared to domain pre-trained models when boosted by task-focused instruction \finetuning.
\end{enumerate}

\section{Related Work}
The field of Large Language Models (LLMs) has seen an avalanche of recent developments. 
GPT3 \citep{brown2020language} took the world by storm with abilities extending beyond what many would consider the ``regular" uses of languages models.
\citet{black2022gpt} presented GPT-NeoX-20B, one of the largest open-source general purpose LLMs, with significant performance on mathematical and knowledge-based tasks. More recently,  Galactica \citep{taylor2022galactica} was one of the major science-based LLMs, and even though it showcased state-of-the-art performance on multiple science-based tasks, the models were retracted due to various technical and ethical issues. Since then, there have been a constant stream of contemporary LLMs \cite{touvron2023llama, rae2022scaling, chowdhery2022palm, hoffmann2022training, thoppilan2022lamda}.

There has also been significant work towards adapting LLMs for a specific domain. \citet{gu2021domain} show that domain-specific pretraining of BERT models serves as a solid foundation for biomedical NLP tasks. SciBERT \citep{beltagy2019scibert}, fine-tuned on domain-specific data,  has state-of-the-art performance on various scientific tasks. There have been several efforts to tune various BERT or BERT-variants for the biomedical domain through continual pretraining~(\citealt{alsentzer2019publicly},~\citealt{peng2019transfer},~\citealt{lee2020biobert},~\citealt{lewis2020pretrained},~\citealt{shin2020biomegatron}) or pretraining from scratch~\citep{gu2021domain}; however, these models are significantly smaller in scale than the current growth of foundation models. In recent times, more works (\citealt{lee2020biobert},\citealt{huang2020clinicalbert}, \citealt{peng2019transfer}) have showcased the benefit of domain-specific training for optimum performance on various domain-specific tasks. 

Likewise, considerable research has been done to achieve domain adaptation using task- or domain-specific instruction \finetuning techniques. \citet{chung2022scaling} show that scaling the number of tasks, model size and finetuning on chain-of-thought data significantly improves the performance across a variety of model classes, prompting setups and evaluation tasks. ALPACA~(\citealt{alpaca}, \citealt{touvron2023llama}, \citealt{wang2022self}), a model fine-tuned from the LLaMA 7B model on 52K instruction-following demonstrations by \citet{alpaca} showcase competitive results to OpenAI's text-davinci-003, while being smaller and cheaper. \citet{ouyang2022training} present InstructGPT, a 1.3B parameter GPT3 model, which sees improvement in general truthfulness and reduction in toxic generation through \finetuning using human-written prompts and reinforcement learning through human feedback. \citet{sanh2022multitask} record improved zero-shot capabilities when a model is fine-tuned on a multitask mixture. Other similar work have shown improved performance through instruction-\finetuned models (\citealt{wei2022finetuned}, \citealt{mishra2022crosstask}, \citealt{aghajanyan-etal-2021-muppet}).

\section{Methods}
In this work, we explore the benefits of pre-training a model with scientific texts and compare this performance to that of general-purpose large language models.
We explore two major generative models as baselines and train them on chemistry-domain scientific literature data, referred to as our in-house AISLE models.
Additionally, after training these models on our scientific data, we  perform instruction \finetuning across the baseline and AISLE models.

\subsection{Data Collection and Processing}
\label{subsec:dataset}

We leverage scientific literature from a variety of data sources when constructing our chemistry-focused pre-training dataset. This included samples from several existing academic literature datasets --- the Semantic Scholar Open Research Corpus (S2ORC)~\citep{lo2019s2orc}, the \textit{Microsoft Academic Graph} (MAG)~\citep{wang2020microsoft}, ArnetMiner's ``AMiner'' dataset~\citep{tang2016aminer}, PubMed publications from \textit{the pile}~\citep{gao2020pile}, the COnnecting REpositories (CORE) dataset~\citep{pontika2016developing}, and the CORD-19 dataset~\citep{wang2020cord} --- and API-based sampling from three publication databases or scholarly search systems: the dblp computer science bibliography (DBLP) ({\url{https://dblp.org}}), Clarivate's Web of Science (WoS), and the Office of Scientific and Technical Information's OSTI.gov engine (OSTI). We also sample publications from two pre-print archives: arXiv~\footnote{\url{https://arxiv.org}} and bioRxiv~\footnote{\url{https://www.biorxiv.org}} as described by \citet{horawalavithana2022foundation}.

In total, our aggregated dataset comprises 53M scientific publication abstracts containing 10B tokens. 

\paragraph{Deduplication}
Recent research has shown that duplicates in training data can significantly affect downstream task performance~\cite{lee2021deduplicating,carlini2022quantifying}. 
In order to detect and remove duplicates, we casefold and strip punctuation from the titles to create a simplified $T'$ and consider two articles $A_{1}$ and $A_{2}$ to be duplicates if they had the same processed title ($T'_{A_{1}} = T'_{A_{2}}$). 

\paragraph{Segmentation and Tokenization}
The large size of our input text made it unfeasible to include the entirety of one text in a single batch for pre-training. However, since truncation of data would result in loss of information and the potential for poorer overall performance~\cite{koh2022empirical}, we chose not to truncate the input data. Instead, apace with others \cite{wang2019multi, dong2023survey}, we concatenate all tokenized strings together and split them into batches, with each batch having a maximum sequence length of 1024. 

Once the documents were segmented, we encoded the text into dense vector embeddings for self-supervised pre-training from scratch. We use the Byte Pair Encoding (BPE) algorithm~\cite{shibata1999byte} to train a tokenizer with a vocabulary size of 64K reflecting the process developed for the GPT2 tokenizer.

\begin{table*}[ht!]
\centering
\small
\begin{tabular}{@{}ll@{}}
\toprule
\textbf{Task} & \multicolumn{1}{c}{\textbf{Instruction Template}} \\
\midrule
CEE & \begin{tabular}[c]{@{}l@{}}Below is an instruction that describes a task. Write a response that appropriately completes the request.\\ \#\#\# Instruction: Identify all \textit{<ENTITY TYPE>} entities in the given text as written.\\ \#\#\# Text: \textit{<INPUT TEXT>}\\ \#\#\# Response: \textit{<LIST OF ENTITIES>}\end{tabular} \\
\midrule
CER & \begin{tabular}[c]{@{}l@{}}Below is an instruction that describes a task. Write a response that appropriately completes the request.\\ \#\#\# Instruction: What are the types of entities in the given text?\\ \#\#\# Text: \textit{<INPUT TEXT>}\\  \#\#\# Response: \textit{<LIST OF ENTITY CLASSES>}\end{tabular}\\
\midrule
MFG & \begin{tabular}[c]{@{}l@{}}Below is an instruction that describes a task. Write a response that appropriately completes the request.\\\#\#\# Instruction: Give the molecular formula for \textit{<IUPAC NAME>}.\\ \#\#\# Response: \textit{<MOLECULAR FORMULA>}\end{tabular}\\
\midrule
ISG & \begin{tabular}[c]{@{}l@{}}Below is an instruction that describes a task. Write a response that appropriately completes the request.\\  \#\#\# Instruction: Give the SELFIE string for \textit{<IUPAC NAME>}.\\ \#\#\# Response: \textit{<SELFIE STRING>} \end{tabular}\\
\midrule
MWE & \begin{tabular}[c]{@{}l@{}}Below is an instruction that describes a task. Write a response that appropriately completes the request.\\\#\#\# Instruction: Give the molecular weight for \textit{<IUPAC NAME>}.\\\#\#\# Response: \textit{<MOLECULAR WEIGHT>}\end{tabular}\\

\bottomrule
\end{tabular}
\caption{\label{tab:instruction_templates}
Prompt templates for the \chemner and PubChem instruction tasks. 
}
\end{table*}

\subsection{Instruction \Finetuning Data}
\label{subsec:instruction_finetuning}

Previous research has shown that the wording and the structure of \finetuning instructions has a large impact on the performance of \finetuned foundation models \cite{wei2022finetuned}.

We used the format of ALPACA 
to create instruction prompts for five chemistry-specific task types using two open-source datasets.
The \chemner~\citep{krallinger2015chemdner} tasks include Chemical Entity Extraction (CEE) and Chemical Entity Recognition (CER). The PubChem \citep{kim2019pubchem} tasks include Molecular Formula Generation (MFG), Isomeric SELFIE String Generation (ISG), and Molecular Weight Estimation (MWE)
\ignore{
\begin{enumerate}
    \item Chemical Entity Extraction (CEE)
    \item Chemical Entity Recognition (CER)
    \item Molecular Formula Generation (MFG)
    \item Isomeric SELFIE String Generation (ISG)
    \item Molecular Weight Estimation (MWE)
\end{enumerate}}
The prompt templates for these tasks are shown in Table \ref{tab:instruction_templates}.

\subsubsection*{\chemner \ Tasks}
We construct instructions for the CER and CEE tasks based on the text, entities and entity classes from the \chemner \ dataset. Each text provided to the model contained one or more chemical named entities from one of seven classes of chemical entities (Trivial, Family, Systematic, Formula, Abbreviation, Multiple, Identifier). For the CEE task, the model had to identify all entities present in the text for a specific entity class. For the CER task, the model had to identify all entity classes for the entities present in the provided text. 

Our custom metrics for CEE and CER tasks based on the methodology described in \citet{chinchor-sundheim-1993-muc} allow us to consider partial matches between the gold and predicted entities. We used four different scoring categories, as described below:
\begin{itemize}
    \item \textit{Correct (COR)}: predicted entities that are exact matches for gold entities.
    \item \textit{Partially Correct (PAR)}: 
    predicted entities with partial overlap with gold entities.
    \item \textit{Incorrect (INC)}: 
    predicted entities that are not present in the gold entities.
    \item \textit{Missing (MIS)}: 
    gold entities that are not present in the list of predicted entities.
\end{itemize}

\noindent
and construct two additional sets: 
\begin{itemize}
    \item Possible Entities:  $POS = COR + PAR + MIS$
    \item Actual Entities: ~~~ $ACT = COR + PAR + INC$
\end{itemize}
to calculate partial scores for precision, recall, and F1  using: 
\begin{align*}
& Precision = (COR + 0.5 * PAR) / (ACT) \\
& Recall = (COR + 0.5 * PAR) / (POS) \\
& F1 = (2 * Prec. * Rec.) / (Prec. + Rec.)
\end{align*} 
    
We also define two evaluation schemas, \textbf{constrained} and \textbf{unconstrained}, indicating whether we constrain the comparison between gold and predicted entities to a specific entity class. For the CEE task, we report results on both constrained and unconstrained schema. For CER, we only report results on the constrained schema.

\subsubsection*{PubChem Tasks}
We constructed instructions for the MFG, ISG, and MWE tasks using the PubChem corpus. For each task, we provided the model with the IUPAC name and instruct it to generate the corresponding molecular formula, generate the isomeric SELFIE string, or estimate the molecular weight. 

\begin{table}[h]
\small
\setlength{\tabcolsep}{3pt} 
\centering
\begin{tabular}{p{0.2\textwidth}p{0.08\textwidth}p{0.08\textwidth}p{0.08\textwidth}}
\toprule
\textbf{Task} & \textit{Train} & \textit{Val} & \textit{Test}\\
\midrule
CEE & 9.6K & 2K & 8.3K \\
CER & 4.5K & 2K & 4.2K\\
MFG, ISG, MWE & 50K & 20K & 5K\\
\midrule
Combined & 164K & 64K & 27.5K\\
\bottomrule
\end{tabular}
\caption{\label{tab:finetune_tasks}
Number of examples for each data split across the five instruction \finetuned tasks. The same molecular data from PubChem was used for the MFG, ISG, and MWE tasks as described in \ref{subsec:instruction_finetuning}.
}
\end{table}

The MFG and ISG tasks are evaluated with the edit distance between the gold and predicted strings based on the algorithm described in \citet{Hyyr2001ExplainingAE}, as implemented in the \textit{editdistance} python package.\footnote{\mbox{\url{https://pypi.org/project/editdistance/}}} We used the sklearn library's implementation of Mean Absolute Percentage Error (MAPE) \footnote{\url{https://scikit-learn.org/stable/modules/generated/sklearn.metrics.mean_absolute_percentage_error.html}} to evaluate performance on the MWE task. We assigned a value of $0.0$ to non-numerical predictions before computing MAPE scores.

\subsection{Models}
In our experiments we compare two core model architectures: the Generative Pre-trained Transformer Model (GPT-2)~\citep{radford2019language} and the BigScience Large Open-science Open-access Multilingual Language Model (BLOOM)~\cite{scao2022bloom}. 
We collect or train the following configurations for each model architecture:
\begin{itemize}
    \setlength\itemsep{-0.5em}
    \item the \offtheshelf baseline model, 
    \item an AISLE model pre-trained from scratch with chemistry-focused data,
    \item a baseline model with instruction \finetuning\!, 
    \item and an AISLE pre-trained from scratch base model with instruction \finetuning\!.
\end{itemize}

\subsubsection*{Baseline Models}
We employ the GPT2-XL model\footnote{\url{https://huggingface.co/gpt2-xl}} with 1.5B parameters and the BLOOM-3B model\footnote{\url{https://huggingface.co/bigscience/bloom-3b}} with 3B parameters for both the first round of analysis contrasting \offtheshelf models with domain pre-trained models as well as for our \finetuning analyses.
We used the pre-trained weights and standard GPT-2 or BLOOM tokenizers available from HuggingFace~\cite{wolf2020huggingfaces}.

\subsubsection*{AISLE Models Trained from Scratch}
We leverage our aggregated scientific data to train seven models from scratch across the two architectures (GPT and BLOOM). We trained these models for three epochs\footnote{Early-stopping was used; models that reached minimum loss before 3 epochs were truncated.} each over 10B tokens from 53 million scientific documents.
All AISLE models were pre-trained with a 95/5 train/validation split.

\subsubsection*{Instruction \Finetuned Models}
\label{subsubsec:instruction-finetuned-models}

In order to better adapt models for the chemistry domain, we perform instruction \finetuning across a variety of tasks. \Finetuning for two epochs with early stopping, we conduct experiments with both the \offtheshelf baseline models and our domain pre-trained AISLE models of each architecture and \finetune on a combination of the training data for all five tasks we consider resulting in four \finetuned models: \baselineGPT, \baselineBLOOM, \aisleGPT, and \aisleBLOOM.

\section{Results}
Next, we discuss the results of three experiments --- each investigating a novel approach to adapting LLMs for domain-specific use. 

\begin{figure*}[ht]
    \centering
    \pgfplotstableread{ 
index	Subset 	{AG0}	{AB0}	{BG0}	{BB0}		{AG3}	{AB3}	{BG3}	{BB3}
0.5	 {Chem\ddag} 	0.288	0.233	0.221	0.228		0.315	0.298	0.226	0.256
1.6	 {ChemBioMed\ddag} 	0.28	0.256	0.238	0.247		0.295	0.273	0.246	0.258
3	 {Health\dag} 	0.281	0.284	0.266	0.265		0.286	0.252	0.255	0.267
4	 {Math} 	0.222	0.199	0.217	0.235		0.257	0.245	0.226	0.266
5	 {STEM\dag} 	0.253	0.244	0.228	0.242		0.27	0.254	0.246	0.265
6	 {Hum.} 	0.26	0.256	0.268	0.257		0.239	0.235	0.259	0.259
7	 {Social Sci.\dag} 	0.278	0.247	0.262	0.273		0.299	0.282	0.264	0.276
8	 {Other\dag} 	0.282	0.289	0.257	0.27		0.277	0.266	0.254	0.271
9	 {Avg\dag} 	0.267	0.259	0.252	0.259		0.271	0.258	0.255	0.268
}\df

  \pgfplotsset{every y tick label/.append style={font=\tiny, xshift=0.5ex}}
    \begin{tikzpicture} 
    \begin{axis}[ 
    ylabel={Accuracy},
    title={\textit{Zero-Shot Accuracy}},
    title style={at={(current bounding box.north west)}, anchor=south west},
title style = {yshift=-0.1in},  
    ylabel style = {yshift=-0.2in}, 
    height=3cm,
    width=16cm, 
    bar width=4,
    axis lines=left,
    xmin=0,
    xmax=9.5,
        ybar,
        ymin=0,
        ymax=0.3,
        xtick=data,
        font=\footnotesize,
        xticklabels from table={\df}{Subset}, 
        legend columns=4,
        legend style={at={(0.5,-0.3)},anchor=north,draw=none,
    column sep=0.75ex,},
    xticklabels={,,,,,,,,},
    ]
    \addplot+[color=baseGPT] table [x=index, meta=Subset, y= BG0] {\df}; 
    \addplot+[color=baseBloom] table [x=index, meta=Subset, y= BB0] {\df}; 
    \addplot+[color=aisleGPT] table [x=index, meta=Subset, y= AG0] {\df}; 
    \addplot+[color=aisleBloom] table [x=index, meta=Subset, y= AB0] {\df}; 

    
    \end{axis}
    \end{tikzpicture}
     \vspace{-.75\baselineskip}

    \begin{tikzpicture} 
    \begin{axis}[ 
    ylabel={Accuracy},
    title={\textit{Few-Shot Accuracy}},
    title style={at={(current bounding box.north west)}, anchor=south west},
title style = {yshift=-0.1in}, 
    ylabel style = {yshift=-0.2in},  
    height=3cm,
    width=16cm, 
    bar width=4,
    axis lines=left,
    xmin=0,
    xmax=9.5,
        ybar,
        ymin=0, 
        ymax=0.3,
        xtick=data,
        font=\footnotesize,
        xticklabels from table={\df}{Subset},
        legend columns=4,
        legend style={at={(0.5,-0.45)},anchor=north,draw=none,
    column sep=0.75ex,}
    ]    
    \addplot+[color=baseGPT,] table [x=index, meta=Subset, y= BG3] {\df}; 
    \addlegendentry{Baseline GPT2-XL}
    \addplot+[color=baseBloom] table [x=index, meta=Subset, y= BB3] {\df}; 
    \addlegendentry{Baseline Bloom-3B}
    \addplot+[color=aisleGPT] table [x=index, meta=Subset, y= AG3] {\df}; 
    \addlegendentry{AISLE GPT2-XL}
    \addplot+[color=aisleBloom] table [x=index, meta=Subset, y= AB3] {\df};
    \addlegendentry{AISLE Bloom-3B}
    \end{axis}
    \end{tikzpicture} 
 
    \caption{Model performance via accuracy in zero-shot (above) and few-shot (below) evaluations across varying subsets of the MMLU benchmark. 
A \ddag denotes both AISLE models outperform baselines across zero and few shot evaluations and \dag indicates at least one AISLE model outperforms both baselines.}
    \label{fig:ht_subsets_plots}
\end{figure*}
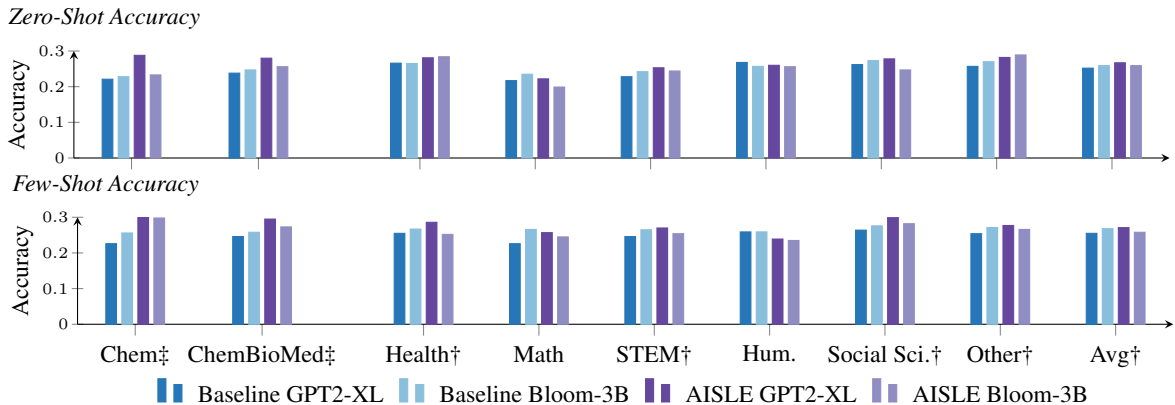

\subsection{Impacts of pre-training from scratch}
First, we analyze the benefits of pre-training with domain-focused datasets for LLM adaptation to a given domain --- \ie addressing the question \textit{what performance gains do trained-from-scratch models offer over \offtheshelf models?} In these experiments we consider both zero-shot and few-shot ($n=3$) evaluations across tasks. 

When we evaluate the model's ability to answer chemistry exam questions at the high school (HT-HC) and college (HT-CC) level, using the chemistry tasks from the MMLU benchmark~\citep{hendrycks2020measuring,hendrycks2021ethics}, we see that our \domainpretrained~ AISLE models outperform \offtheshelf baseline models, as shown in  \autoref{tab:pretrain_performance_comparison_acc} that illustrates consistent strong performance by the \aisleGPT model.

When we evaluate the models across the MMLU benchmark as a whole, comparing average performance across all tasks regardless of topic, we also see that our strongest AISLE model (\aisleGPT) outperforms both \offtheshelf models by a small margin. \autoref{fig:ht_subsets_plots} illustrates our performance comparisons across a variety of topic subsets\footnote{Chemistry: [college chemistry, high school chemistry],
 ChemBioMed: [college chemistry, high school chemistry, college biology, high school biology, clinical knowledge, college medicine, medical genetics, professional medicine, virology],
 Mathematics: [abstract algebra, college mathematics, elementary mathematics, high school mathematics, high school statistics],
 Health: [anatomy, clinical knowledge, college medicine, human aging, medical genetics, nutrition, professional medicine, virology],
 STEM: [college chemistry, high school chemistry, astronomy, college physics, high school physics, conceptual physics, college biology, high school biology, college computer science, high school computer science, computer security, machine learning, electrical engineering, abstract algebra, college mathematics, high school mathematics, high school statistics, elementary mathematics],
 Humanities: [high school european history, high school us history, high school world history, prehistory, formal logic, logical fallacies, moral disputes, moral scenarios, philosophy, world religions, international law, jurisprudence, professional law],
 Social Sciences: [high school government and politics, public relations, security studies, us foreign policy, human sexuality, sociology, econometrics, high school macroeconomics, high school microeconomics, high school geography, high school psychology, professional psychology],
 Other: [global facts, miscellaneous, business ethics, professional accounting, management, marketing, anatomy, clinical knowledge, college medicine, human aging, medical genetics, nutrition, professional medicine, virology]}, where \ddag~ denotes topics where both AISLE models outperform baselines across both evaluations and \dag~ indicates at least one AISLE model outperforms both baselines.
We find that a \domainpretrained~ AISLE model outperforms consistently in Chemistry focused topics (Chem, ChemBioMed) and Chemistry-adjacent topics (Health, STEM). Interestingly, we see some higher performance compared to baselines in the Social Science (Social Sci.) and ``Other'' (\eg business, accounting) focused topics from MMLU. 

\begin{table}[tb]
\small
\centering
\setlength{\tabcolsep}{3pt} 
\textit{MMLU Accuracy}
\begin{tabular}{lrrrr}
\toprule
{} &     \multicolumn{2}{c}{HT-HC} &  \multicolumn{2}{c}{HT-CC} \\
&     {\footnotesize 0-shot} &  {\footnotesize 3-shot} &     {\footnotesize 0-shot} & {\footnotesize 3-shot} \\
\midrule
\aisleGPT                &  \textbf{0.27} &  \textbf{0.27}&   \textbf{0.31}&   \textbf{0.36} \\ 
 \aisleBLOOM               &  0.21 &  {0.25} &   {0.26} &   \textbf{0.35} \\ 
\midrule
 \baselineGPT           &  0.18 &  0.20&   0.26&   0.25 \\ 
 \baselineBLOOM          &  {0.23} &  0.23 &   0.23  &   0.28 \\ 
 \midrule
 Random Baseline                    & 0.25 &  0.25 &  0.25 &   0.25 \\
\bottomrule
\end{tabular}

\caption{\label{tab:pretrain_performance_comparison_acc}
MMLU~ High School Chemistry (HT-HC) and College Chemistry (HT-CC) performance comparing zero- and three-shot settings --- best performance across settings is highlighted in bold.}
\end{table}

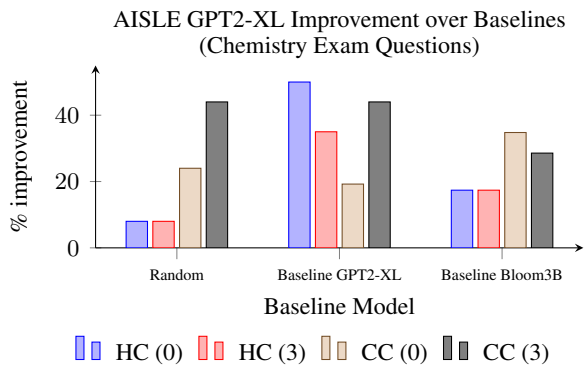
\begin{figure}[t!]
    \centering
    \pgfplotstableread{
index      Baseline    HTHC0 HTHC3 HTCC0 HTCC3 
1     {\tiny Random}  8	8	24	44
2       {\tiny {Baseline}~GPT2-XL}  50   35	19.23	44
3       {\tiny {Baseline} Bloom3B}  17.39	17.39	34.78	28.57
}\df

    \begin{tikzpicture} 
    \begin{axis}[ 
    xlabel=Baseline Model,
    ylabel={\% improvement},
    title={\footnotesize AISLE GPT2-XL Improvement over Baselines \\ (Chemistry Exam Questions)},
    title style = {align = center},
title style = {yshift=-0.1in}, 
ylabel style = {yshift=-0.1in}, 
    height=4cm,
    width=8cm, 
    bar width=8,
    axis lines=left,
    xmin=0.5,xmax=3.5,
        ybar,
        ymin=0,
        ymax=55,
        xtick=data,
        font=\footnotesize,
        xticklabels from table={\df}{Baseline},
        legend columns=-1,
        legend style={at={(0.45,-0.45)},anchor=north,draw=none,
    column sep=0.75ex,}
    ]
    \addplot+ table [x=index, meta=Baseline, y= HTHC0] {\df}; 
    \addlegendentry{HC (0)}
    \addplot+ table [x=index, meta=Baseline, y= HTHC3] {\df}; 
    \addlegendentry{HC (3)}
    \addplot+ table [x=index, meta=Baseline, y= HTCC0] {\df}; 
    \addlegendentry{CC (0)}
    \addplot+ table [x=index, meta=Baseline, y= HTCC3] {\df}; 
    \addlegendentry{CC (3)}
    \end{axis}
    \end{tikzpicture}
     \vspace{-1.5\baselineskip}
    \caption{Relative improvement in accuracy, ranging from 10\%-50\%, achieved by the \aisleGPT model over baselines for zero-shot (0) and few-shot using 3 examples (3).}
    \label{fig:relImprovementPretrainBase}
    \vspace{-0.75em}
\end{figure}

Focusing specifically on the two chemistry exam tasks in the MMLU benchmark individually, we find that our pre-trained GPT2 and BLOOM models outperform their respective \offtheshelf variants in both the zero-shot and few-shot settings. We see this across both chemistry tasks and we find consistently strong performance by our \aisleGPT. Comparisons of this model's relative improvement over baseline models, as illustrated in \autoref{fig:relImprovementPretrainBase}, shows particular improvement over \offtheshelf models for the college-level questions. As would be expected, the few-shot setting consistently matches or outperforms the zero-shot setting for all baseline and domain pre-trained models except for the \baselineGPT model on college-level questions where we see a slight dip in performance ($-0.01$). Our results demonstrate that training from scratch on domain-focused data significantly improves performance.

We also evaluated this suite of models on out-of-domain perplexity-based tasks (Open AI LAMBADA~\citep{paperno2016lambada}, PubMed Abstracts from the Pile~\citep{gao2020pile}, and Wikitext~\citep{merity2016pointer}. However, the results indicated that while pre-training from scratch may be beneficial for in-domain tasks, there is not enough change or improvement to draw conclusions about their impact on out-of-domain tasks: performance increased on the Pile PubMed Abstracts \citep{gao2020pile} task and decreased on the LAMBADA and Wikitext. However, we note that the two with decreased results are further ``out of domain'' than the PubMed Abstract data sample. 

\subsection{Impacts of Instruction \Finetuning}
For our next two analyses, we examine the performance of \offtheshelf Baseline models and pre-trained AISLE models to identify what benefit instruction \finetuning brings to chemistry-focused extractive and generative tasks. 
First, we evaluate on held-out data from the five tasks incorporated during the instruction \finetuning and then we examine the impacts instruction \finetuning can have on unseen, in-domain multiple choice tasks (\ie MMLU) described in the previous section. 

\subsubsection*{Evaluation on Instruction Tasks}
We present the performance of models on instruction \finetuning tasks in \autoref{tab:instruction_task_performance_comparison_inst_f1_BLonly}. 
When we evaluate performance of the models on \chemner~ tasks, we see that our from-scratch trained model with added instruction fine-tuning \aisleGPT surpasses the performance of the baseline models (i.e., those that have only been instruction \finetuned without previous training) on the CER task. For CEE, 
the \aisleGPT models are very close in performance to Baseline models. As expected, the performance on unconstrained schema for CEE tasks is consistently higher for all the models than the constrained schema.

\begin{figure*}[ht]
  \centering
\begin{tabular}{cc}
\includegraphics[width=0.48\linewidth]{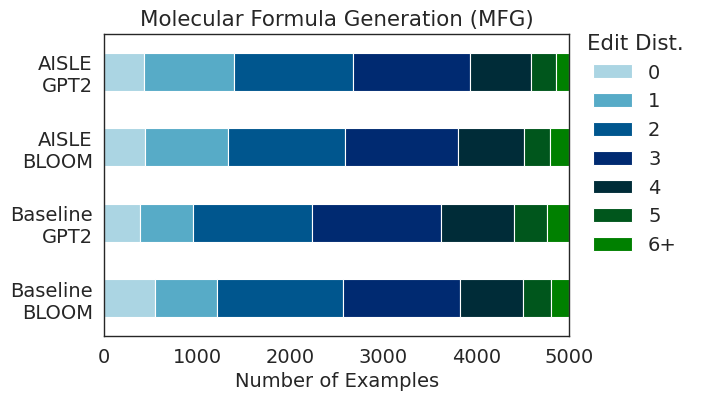} 
&
\includegraphics[width=0.5\linewidth]{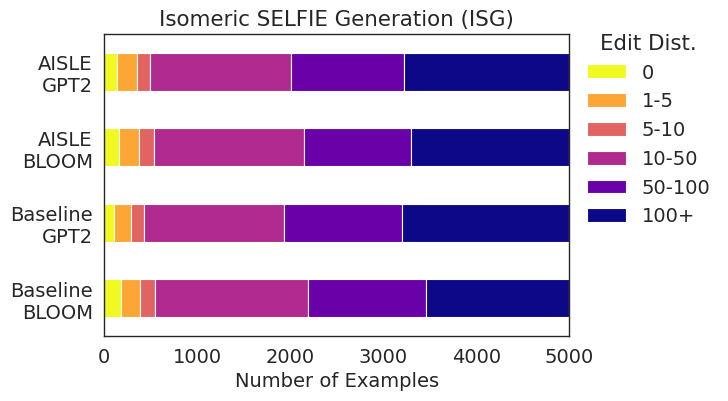}
\\
\includegraphics[width=0.44\linewidth]{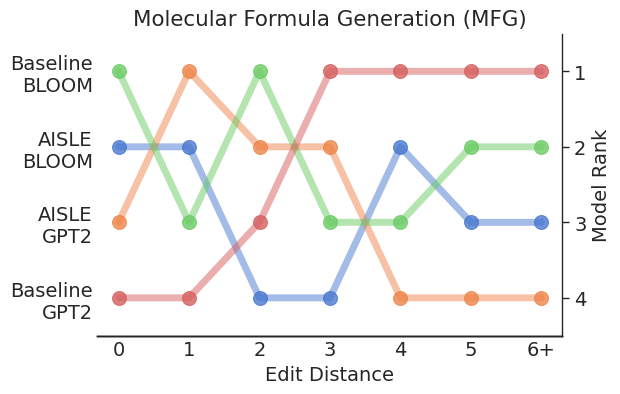}
  \hspace{6mm}
&
\includegraphics[width=0.44\linewidth]{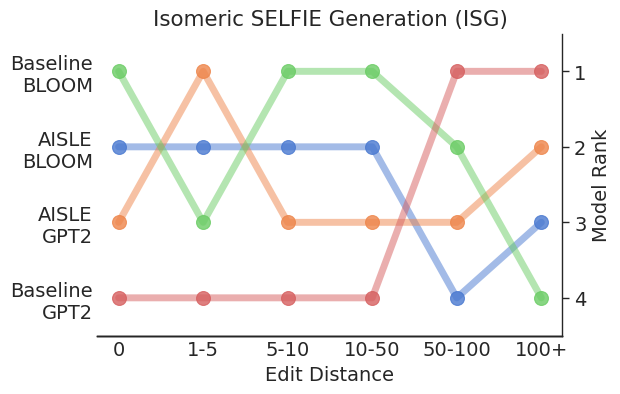}
  \hspace{6mm}
\end{tabular}

  \vspace{-1.5em}
  \caption{Bar plots show the number of examples with each edit distance per model. Bump charts rank the models based on the number of examples with each edit distance -- we ideally want to see a rank of 1 when edit distance is lower and a rank of 4 when edit distance is higher. }
  \label{fig:generationplots}
\end{figure*}

\aisleGPT models also surpass performance of Baseline models for the MFG and MWE PubChem tasks. 
The variability in edit distance across models for MFG and ISG is displayed in \autoref{fig:generationplots} where we see that both AISLE models surpass that of the \baselineGPT. For the MFG task, we see that the AISLE models illustrate stronger performance at low edit distances (in particular perfect or near perfect [0,1]) over both baselines although \baselineBLOOM shows slightly higher numbers for edit distance of 0.

\subsubsection*{Evaluation on Multiple Choice Tasks}
Across the architectures, we find that the AISLE models trained from scratch on in-domain data with few-shot prompting outperform Baseline \offtheshelf models indicating that while instruction \finetuning can provide a reasonable performance improvement for domain adaptation, it still lags behind models trained fully from scratch. However, because training models from scratch can acrue high cost in terms of time, money, and compute, instruction \finetuning can offer a potential alternative for low-resource use cases.

\begin{table}[ht]
\small
\centering
\setlength{\tabcolsep}{3pt}

\begin{tabular}{lllr}
\toprule
& \multicolumn{2}{l}{\textit{Const. F1}} & \textit{Unconst. F1} \\
\textbf{\chemner ~ Tasks}
&         CER &    CEE &           CEE \\
\midrule
\aisleGPT       &       \textbf{0.816} &  0.427 &         0.493 \\
\aisleBLOOM      &       0.433 &  0.285 &         0.304 \\
\midrule
\baselineGPT  &       0.787 &  \textbf{0.433} &         \textbf{0.507} \\
\baselineBLOOM &       0.813 &  0.407 &         0.479 \\
\bottomrule
\end{tabular}

\vspace{0.25\baselineskip}

\begin{tabular}{lllr} 
\toprule
& \multicolumn{2}{l}{\textit{Mean \% Edit Dist.}} &   \textit{MAPE} \\
\textbf{PubChem Tasks~~~~~~}
&                ISG &    MFG &    MWE \\
\midrule
\aisleGPT       &              0.395 &  \textbf{0.239} &  \textbf{0.063} \\
\aisleBLOOM      &              0.371 &  0.246 &  \textbf{0.063} \\
\midrule
\baselineGPT  &              0.409 &  0.272 &  0.090 \\
\baselineBLOOM &              \textbf{0.355} &  0.247 &  0.152 \\
\bottomrule
\end{tabular}

\caption{\label{tab:instruction_task_performance_comparison_inst_f1_BLonly}Performance on the instruction tasks for the instruction \finetuned models, with best performance for each task highlighted in bold.}
\end{table}
\begin{table}[t]
\small
\centering
\setlength{\tabcolsep}{5pt} 
\begin{tabular}{lcccc}
\multicolumn{5}{c}{\textit{Zero-Shot Evaluation} }\\
\toprule
{} & \multicolumn{2}{c}{High School} & \multicolumn{2}{c}{College} \\
{} &      Base & +Instr. &     Base & +Instr. \\
\midrule
\aisleGPT       &     \textbf{0.268} &   0.218 &     \textbf{0.280} &   0.315 \\
\aisleBLOOM      &     0.194 &   0.227 &     0.233 &   \textbf{0.341} \\
\midrule
\baselineGPT  &     0.162 &   0.229 &     0.230 &   0.246  \\
\baselineBLOOM &     0.218 &   \textbf{0.246} &     0.224 &   0.279\\
\bottomrule
\\
\end{tabular}

\begin{tabular}{lcccc}
\multicolumn{5}{c}{\textit{Few-Shot Evaluation} }\\
\toprule
{} & \multicolumn{2}{c}{High School} & \multicolumn{2}{c}{College} \\
{} &      Base & +Instr. &     Base & +Instr. \\
\midrule

\aisleGPT       &    \textbf{0.262} &   0.221 &    0.303 &   \textbf{0.332} \\
\aisleBLOOM      &    0.230 &   \textbf{0.275} &    \textbf{0.354} &   0.274 \\
\midrule
\baselineGPT  &    0.196 &   0.206&    0.235 &   0.293 \\
\baselineBLOOM &    0.232 &   0.252 &    0.278 &   0.327 \\
\bottomrule
\end{tabular}

\caption{\label{tab:ht_performance_all_comparison}
{Performance} on zero-shot prompting for the High School Chemistry and College Chemistry MMLU tasks using Macro F1. Base pre-trained models (Base) are compared to the same base after instruction \finetuning on tasks listed in Table \ref{tab:finetune_tasks} (+Instr.). Best performance is highlighted in bold.
}
\vspace{-0.75em}
\end{table}

In \autoref{tab:ht_performance_all_comparison}, we highlight the performance of instruction \finetuned models on held-out tasks.  When we consider the unseen MMLU high school- and college-level chemistry multiple choice tasks, we find that the  performance on the BLOOM models across the zero-shot and few-shot prompting increased significantly after instruction fine-tuning. 
This is a good indication that adding instruction \finetuning can lead to better performance on novel, in-domain tasks, both on its own and as an added training on top of training from scratch. However, when we compare the instruction \finetuned \aisleBLOOM and \baselineBLOOM models, the results do not hold a clear pattern -- with zero-shot prompting we see a drop in performance whereas in the few-shot setting we see an increase in performance.

\section{Discussion}
\label{sec:discussion}

In this work we provide an exploration of performance trade-offs for three key methods of domain-adaption. Our results have highlighted some of the trade-off space for performance in compute (\eg resource-constrained environments that may inhibit the use of an extremely large LLM) or API-access constrained environments. We highlight the strength of domain pre-training from scratch to support varied unseen tasks, with three key findings that we highlight to prompt future research. 

First, we see that while the performance of the BLOOM models increases for perplexity-based tasks when instruction \finetuned on the chemistry tasks, GPT models don't exhibit such consistent results. Their performance varies with seemingly no pattern as to why this occurs. One possible explanation is because the instruction \finetuning set contained molecular formula and SELFIE strings, which are different (structurally and semantically) from regular written prose tokens, causing the model to produce much less coherent text when asked to generate regular words. But this does not explain why this effect is not seen when using BLOOM models (off-the-shelf, or using domain-focused text), warranting further study.

We also highlight the poor performance of our instruction \finetuned models on the PubChem tasks, in contrast to much better performance on the ChemDNER tasks. We believe this is because the tasks require the models to generate or interpret molecular formulae again suffer from the aforementioned fact that the patterns for those are dissimilar from scientific writing text (\eg similar to out of vocab or unseen languages).

Finally, we see that for the BLOOM models, the performance actually decreases for the instruction-based CHEMDNER and PubChem tasks when instruction \finetuned on these tasks. In several instances, the model simply generates an empty string as the output for the queries. We have not conducted any study into why this might have occurred beyond replicating training and evaluation to confirm it is consistent with GPT2-XL model evaluation, and leave it as a future direction.

\section{Limitations}
The experiments in this work have explored in detail the impacts of the three key methods to adapt models for the scientific domains. During these experiments, one main factor of limitation was the computational cost of training large models from scratch. Given more time and resources, we would have liked to train our models for greater number of epochs than our current limit of three. Additionally, the sparsity of appropriate benchmarks for LLMs in the science domain led us to adapt existing benchmarks for evaluation that were not designed for this specific purpose.

In the future, this work can be expanded such that the training data includes not just articles, but a varied form of data. It would also be interesting to see how several other new LLMs perform in our experimental settings. Finally, it would be a good study to see if retriever-augmented language models can outperform these performances given the same training data.

\section{Ethical Consideration}
It has generally been the norm to assume that previously published work can be used as-is without having to consider the inherited ethical issues. However, in present  times, researchers should not “simply assume that [...] research will have a net positive impact on the world” \cite{hecht2021s}. We acknowledge that this applies not just to new work, but also when using existing work in the way that we have done.

While we do not anticipate the novel work presented here to introduce new ethical concerns in and by themselves, we do recognize that there may also be pre-existing concerns and issues of the data, models, and methodologies we have used for this paper. In particular, it has been seen that Large Language Models (LLMs), like the ones used in this work, exhibit a wide variety of bias -- \eg religious, gender, race, profession, and cultural -- and frequently generate answers that are incorrect, misogynistic, antisemitic, and generally toxic. \cite{abid2021persistent,buolamwini2018gender,liang2021towards,nadeem-etal-2021-stereoset,welbl2021challenges}
However, when used within the parameters of our experiments detailed in this paper, we did not see such behaviour from any of the models. 
To our knowledge, when used as intended, our models do not pose additional ethical concerns than any other LLM.

\section{Acknowledgements}
The research described in this paper is part of the MARS Initiative at Pacific Northwest National Laboratory, which is operated by Battelle Memorial Institute for the U.S. Department of Energy under contract DE-AC05-76RLO1830. Any opinions, findings, and conclusions or recommendations expressed in this material are those of the author(s) and do not necessarily reflect the views of the United States Government or any agency thereof. This article has been cleared by PNNL for public release as PNNL-SA-191592.

\bibliography{references}

\end{document}